\documentclass[10pt,twocolumn,letterpaper]{article}

\usepackage{cvpr}              

\newtheorem{theorem}{Theorem}[section]

\newtheorem{assumption}[theorem]{Assumption}
\newtheorem{remark}[theorem]{Remark}

\definecolor{cvprblue}{rgb}{0.21,0.49,0.74}
\usepackage[pagebackref,breaklinks,colorlinks,allcolors=cvprblue]{hyperref}

\def\paperID{721} 
\def\confName{CVPR}
\def\confYear{2026}

\title{Breaking the Modality Wall: Time-step Mixup for\\Efficient Spiking Knowledge Transfer from Static to Event Domain}

\author{
Yuqi Xie\textsuperscript{1}\thanks{Equal contribution.} \quad
Shuhan Ye\textsuperscript{1,2}\footnotemark[1] \quad
Yi Yu\textsuperscript{2}\thanks{Corresponding authors: wangchong@nbu.edu.cn, yu.yi@ntu.edu.sg} \quad
Chong Wang\textsuperscript{1,3}\footnotemark[2] \quad
Qixin Zhang\textsuperscript{2} \\
Jiazhen Xu\textsuperscript{1} \quad
Le Shen\textsuperscript{1} \quad
Yuanbin Qian\textsuperscript{1} \quad
Jiangbo Qian\textsuperscript{1,3} \quad
Guoqi Li\textsuperscript{4}\\[0.3em]
\textsuperscript{1}Ningbo University \quad
\textsuperscript{2}Nanyang Technological University\\
\textsuperscript{3}Merchants' Guild Economics and Cultural \quad
\textsuperscript{4}Institute of Automation, Chinese Academy of Sciences\\[0.3em]
{\tt\small 2411100305@nbu.edu.cn, shuhan006@e.ntu.edu.sg, yu.yi@ntu.edu.sg}\\
{\tt\small wangchong@nbu.edu.cn, qixin.zhang@ntu.edu.sg}\\
{\tt\small \{2311100314, 2411100289, 2311100301, qianjiangbo\}@nbu.edu.cn,
guoqi.li@ia.ac.cn}
}

\begin{document}
\maketitle
\begin{abstract}

The integration of event cameras and spiking neural networks (SNNs) promises energy-efficient visual intelligence, yet scarce event data and the sparsity of DVS outputs hinder effective training. Prior knowledge transfers from RGB to DVS often underperform because the distribution gap between modalities is substantial. In this work, we present \textbf{Time-step Mixup Knowledge Transfer (TMKT)}, a cross-modal training framework with a probabilistic \textbf{Time-step Mixup (TSM)} strategy. TSM exploits the asynchronous nature of SNNs by interpolating RGB and DVS inputs at various time steps to produce a smooth curriculum within each sequence, which reduces gradient variance and stabilizes optimization with theoretical analysis. To employ auxiliary supervision from TSM, TMKT introduces two lightweight modality-aware objectives, \textit{Modality Aware Guidance (MAG)} for per-frame source supervision and \textit{Mixup Ratio Perception (MRP)} for sequence-level mix ratio estimation, which explicitly align temporal features with the mixing schedule. TMKT enables smoother knowledge transfer, helps mitigate modality mismatch during training, and achieves superior performance in spiking image classification tasks. Extensive experiments across diverse benchmarks and multiple SNN backbones, together with ablations, demonstrate the effectiveness of our method.

\end{abstract}    
\section{Introduction}
In recent years, the integration of event cameras and spiking neural networks (SNNs) has attracted significant attention~\cite{merolla2014truenorth,davies2018loihi,roy,li2021dspike,wang2023asgl,zheng2021tdbn,kudithipudi2025neuromorphic}. Event cameras, also known as dynamic vision sensors (DVS)~\cite{lenero20113}, are inspired by the mammalian brain and capture visual data in response to changes in light intensity. This makes them an ideal solution for addressing the limitations of conventional cameras, such as low dynamic range and frame rate \cite{atis, celex, davis}. Meanwhile, SNNs are inherently well-suited for processing event-driven inputs while offering impressive energy efficiency. Their ability to process temporal information aligns perfectly with the high temporal resolution provided by event cameras \cite{deng2021}. The synergy between these two bio-inspired technologies presents a compelling approach for tackling low-power vision tasks like image classification \cite{spikformer,ckd}, action recognition \cite{sdv3}, and video anomaly detection \cite{ucfcrimedvs}.

\begin{figure}[t]
\centering
\includegraphics[width=1.0\columnwidth]{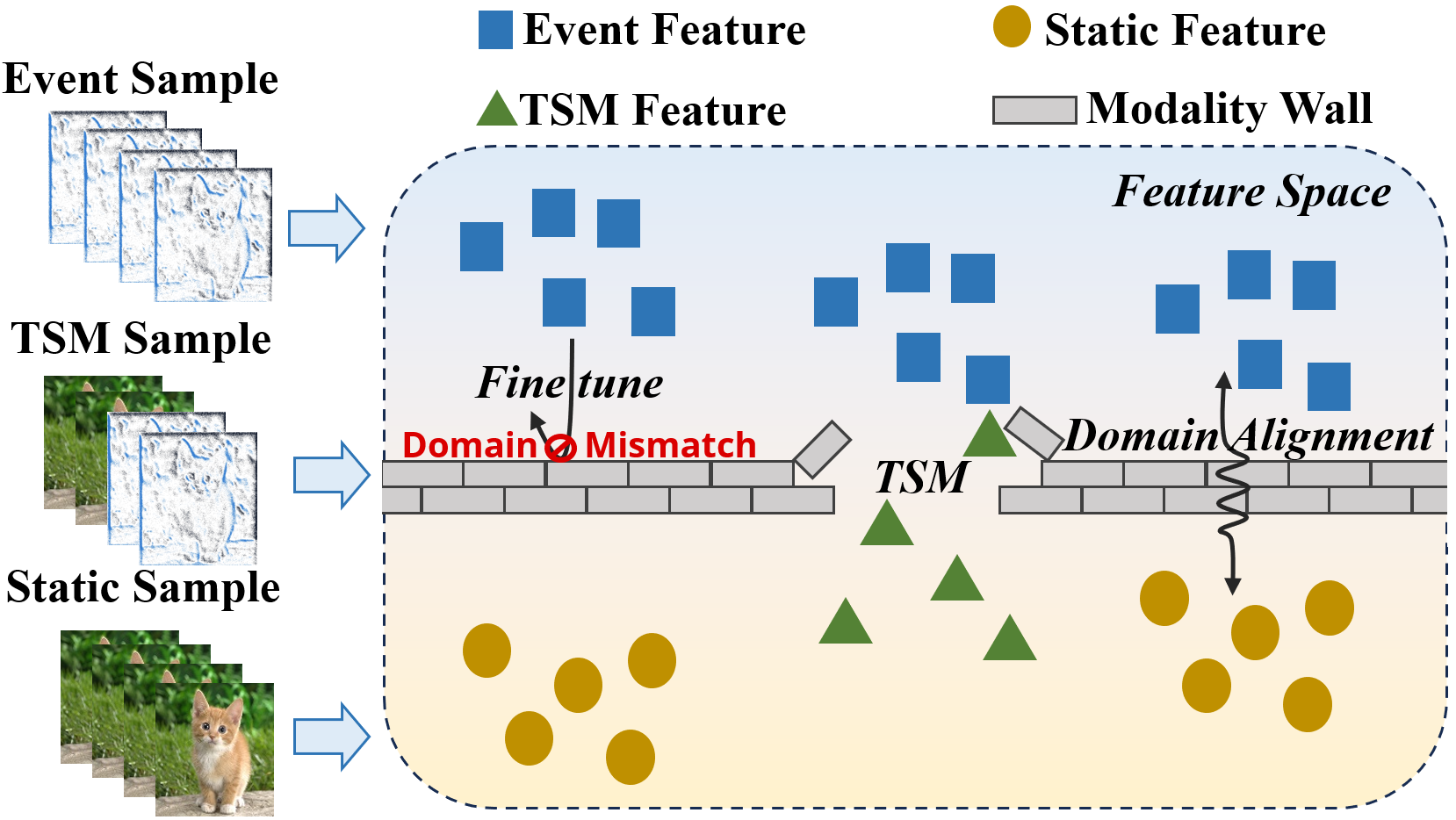}
\vspace{-6mm}
\caption{Different paradigms for transferring static (RGB) knowledge to the event (DVS) domain.Finetuning on DVS data suffers from severe domain mismatch, causing the representations to hit the modality wall and fail to cross domains; Domain alignment alleviates this mismatch by explicitly pulling the DVS representations toward the RGB manifold; Our proposed Time-step Mixup (TSM) provides a smoother, causality-aligned transition across the modality wall, enabling stable cross-domain learning in the shared feature space.}
\label{fig1}
\vspace{-2mm}
\end{figure}


Despite their advantages, the application of event cameras faces significant challenges, due to the costly and time-consuming data acquisition process~\cite{zhu2018multivehicle,gehrig2021dsec,gehrig2020video}. This results in small-scale and hard-to-access datasets that hinder its further development. Additionally, since DVS only capture and encode brightness changes exceeding a certain threshold, they are primarily sensitive to the edges of moving objects~\cite{gallego2020event}. This means they discard a substantial amount of contextual cues, such as background, texture and color, which are crucial for comprehensive semantic understanding needed in high-level vision tasks~\cite{rebecq2019events}.
In contrast, static datasets (RGB ones) offer abundant contextual information, and are both large-scale and easily accessible. 
These datasets provide richer details that support more robust analysis across various applications~\cite{deng2009imagenet}.


However, due to the significant distribution gap between static and event domains, as shown in Fig. \ref{fig1}, directly fine-tuning models pre-trained on static datasets often leads to negative transfer. To address this, several studies explore knowledge transfer that bridges modalities by moving rich semantics from static to event \cite{stl, ekt}. Yet these methods typically overlook the raw-level discrepancy: RGB frames are dense and intensity-rich, while DVS frames are highly sparse, with values clustered near zero and markedly different dynamic ranges. A recent Knowledge-Transfe approach \cite{ekt} tackles distribution shift with a sliding data replacement strategy, gradually substituting source (static) samples with target (event) samples within the batch so the model transitions from source-dominated to target-dominated under a controlled ratio. Nevertheless, this process implicitly assumes similar feature distributions across modalities, which rarely holds in practice. Consequently, when dissimilar modalities coexist within the same batch, intra-batch modality shifts arise and complicate learning.

Inspired by the concept of such data replacement  \cite{ekt}, we aim to develop a smoother and more refined approach for cross-modal data mixing that guides the model to effectively handle heterogeneous modalities. 
A natural starting point is to examine successful MixUp-like strategies \cite{mixup,cutmix}, which provide valuable insights through their approach of spatially blending inputs and labels simultaneously. However, these techniques are primarily designed for intra-class interpolation within a single modality. The substantial gap between RGB and DVS data presents challenges that make direct input-level and label interpolation impractical in our scenario. 



To address these challenges, we present \textbf{Time-step Mixup Knowledge Transfer (TMKT)}, a cross-modal training framework built around a \textbf{probabilistic Time-step Mixup (TSM)} strategy tailored for SNNs. 
TSM performs probabilistic time-step mixup by truncating portions of the apperance (RGB) sequence and stochastically replacing selected steps with event inputs, forming a smooth within-sequence curriculum that is compatible with asynchronous spiking computation.
From our theoretical analysis, such TSM reduces gradient variance and stabilizes optimization, facilitating smoother and more gradual knowledge transfer as shown in Fig.~\ref{fig1}. To provide auxiliary supervision and reduce temporal ambiguity, TMKT further introduces two lightweight modality-aware objectives: \textit{Modality Aware Guidance (MAG)} for per-frame source supervision and \textit{Mixup Ratio Perception (MRP)} for sequence-level mix ratio estimation, which encourage temporal features to follow the mixing schedule. Together, these components mitigate modality gap and yield consistent improvements on spiking image classification.

Our contribution can be summarized as follow:


\begin{itemize}
\item We introduce \textbf{TMKT}, a cross-modal framework grounded in a \emph{probabilistic} Time-step Mixup (TSM). By truncating the RGB sequence and stochastically substituting chosen time steps with event inputs, TSM produces a smooth temporal transition consistent with spiking computation with theoretical analysis. To the best of our knowledge, TMKT constitutes the first mixup perceptive strategy for spiking knowledge transfer.

\item Cross-modal TSM provides auxiliary supervisions via two lightweight objectives: \emph{Modality Aware Guidance (MAG)} for per-frame source, and \emph{Mixup Ratio Perception (MRP)} for sequence-level mix ratio, to keep temporal features aligned with the mixing schedule. 

\item Extensive evaluations across diverse SNN backbones and benchmarks, along with extensive ablation studies, substantiate consistent gains of our TMKT.
\end{itemize}

\section{Related Work}
\paragraph{Spiking Neuron Models.}
SNNs draw inspiration from the human brain, using discrete spikes for information processing. This method achieves effects comparable to continuous activation functions by accumulating spikes over an additional temporal dimension, making it highly suitable for processing temporal data. Concretely, SNNs replace the traditional activation function by using a spiking neuron model, such as the Integrate-and-Fire (IF) neuron model \cite{IF}  and the widely-used Leaky Integrate-and-Fire (LIF) neuron model \cite{LIF}. The LIF neuron model integrates incoming spikes over time, with its membrane potential and spiking behavior governed by the following equations:
\begin{equation}
\begin{split}
\hat{\textbf{u}}^{t,l} &= \tau\,\textbf{u}^{t-1,l} + \textbf{W}^{l}\textbf{s}^{t,l-1},\\
    \textbf{s}^{t,l} &= H\!\big(\hat{\textbf{u}}^{t,l} - V_{\text{th}}\big),
    ~{\textbf{u}}^{t,l} = \hat{\textbf{u}}^{t,l}\big(1 - \textbf{s}^{t,l}\big).
\end{split}
\label{eq2}
\end{equation}
where \( \hat{\textbf{u}}^{t,l} \) is the updated membrane potential of neurons in layer \( l \) at time-step \( t \), \( \textbf{u}^{t,l} \) is the membrane potential of layer \( l \) after a spike at time-step \( t \), \( \textbf{W}^{l} \) represents the weight matrix of layer \( l \), and \( \textbf{s}^{t,l} \) corresponds to the binary spikes emitted by neurons. The Heaviside step function \( H \) determines whether a spike is emitted, based on the comparison between \( \textbf{u}^{t,l} \) and the threshold \( V_{\text{th}} \). The leaky factor \( \tau \) controls the temporal decay of the membrane potential. 

\vspace{1mm}
\noindent\textbf{Spiking Knowledge Transfer.}
Knowledge transfer has been widely applied in traditional ANNs and has achieved remarkable success \cite{wang2019domain}. 
However, in SNNs which have attracted increasing attention due to their energy efficiency, transfer learning remains relatively underexplored.
In particular, transferring knowledge from static domains (\textit{e.g.,} grayscale or color images) to the event domain (e.g., DVS) using SNNs holds great potential for addressing the challenges of limited scale and accessibility of DVS datasets, which often lead to poor generalization performance.
Although research in this area remains limited, a few recent works have started to explore knowledge transfer from static to event domains.
Specifically, R2ETL~\cite{stl} utilizes labeled RGB data for SNN transfer learning by introducing encoding and feature alignment modules, and extends CKA to TCKA. EKT~\cite{ekt} proposes a gradual replacement strategy, where static images are progressively replaced by event data during training, guided by a loss combining domain alignment and spatio-temporal regularization. CKD~\cite{ckd} adopts phased cross-architecture distillation, transferring static-domain features from ANNs to SNNs. 
These methods rely on shared-parameter backbones but overlook distribution gaps between source and input domains, leading to suboptimal transfer. 
We propose a smoother transfer approach, with a modality-aware guidance that mitigates the issues caused by input data distribution differences.

\vspace{1mm}
\noindent\textbf{Mixup for Cross-Modality Transfer.}
Many works~\cite{guo2019mixup,hu2021neural,wang2022vlmixer} leverage MixUp-style data augmentation to bridge modality gaps. AdaMixUp \cite{guo2019mixup} views MixUp as a form of out-of-manifold regularization and addresses its limitations via adaptive mixing strategies. Neural Dubber \cite{hu2021neural} introduces a multi-modal TTS system that synchronizes speech with video using lip movements and speaker embeddings. VLMixer \cite{wang2022vlmixer} applies cross-modal CutMix for unpaired vision-language pretraining, achieving effective alignment between image and text modalities. Despite their success in other domains, MixUp-style strategies remain unexplored for transfer learning in SNNs. Transferring from static to event domains is challenging due to large intensity distribution gaps, which make direct interpolation ambiguous. Notably, SNNs process static inputs by repeating frames across time-step and averaging outputs. Leveraging this structure, we propose a temporal replacement strategy that substitutes entire frames along the time-step axis, preserving semantics while enabling smoother cross-domain transfer.
\section{Methodology}
\vspace{-2mm}
In this section, we present a new SNN-based framework, namely Time-step Mixup Knowledge Transfer (TMKT) model, designed to transfer knowledge from the static domain to the event domain. 
Specifically, TMKT integrates three key components: the Time-step Mixup (TSM) strategy, the Modality-Aware Guidance (MAG), and the Mixup ratio Perception (MRP).
By leveraging the inherent temporal heterogeneity of SNNs, TMKT constructs a robust common representation space between the source and target domains. This facilitates bridging the modality-induced domain gap and enables smooth and efficient transfer knowledge of source domain to the target domain.

\subsection{Overall Architecture}
\begin{figure*}[!t]
\centering
\includegraphics[width=1.0\textwidth]{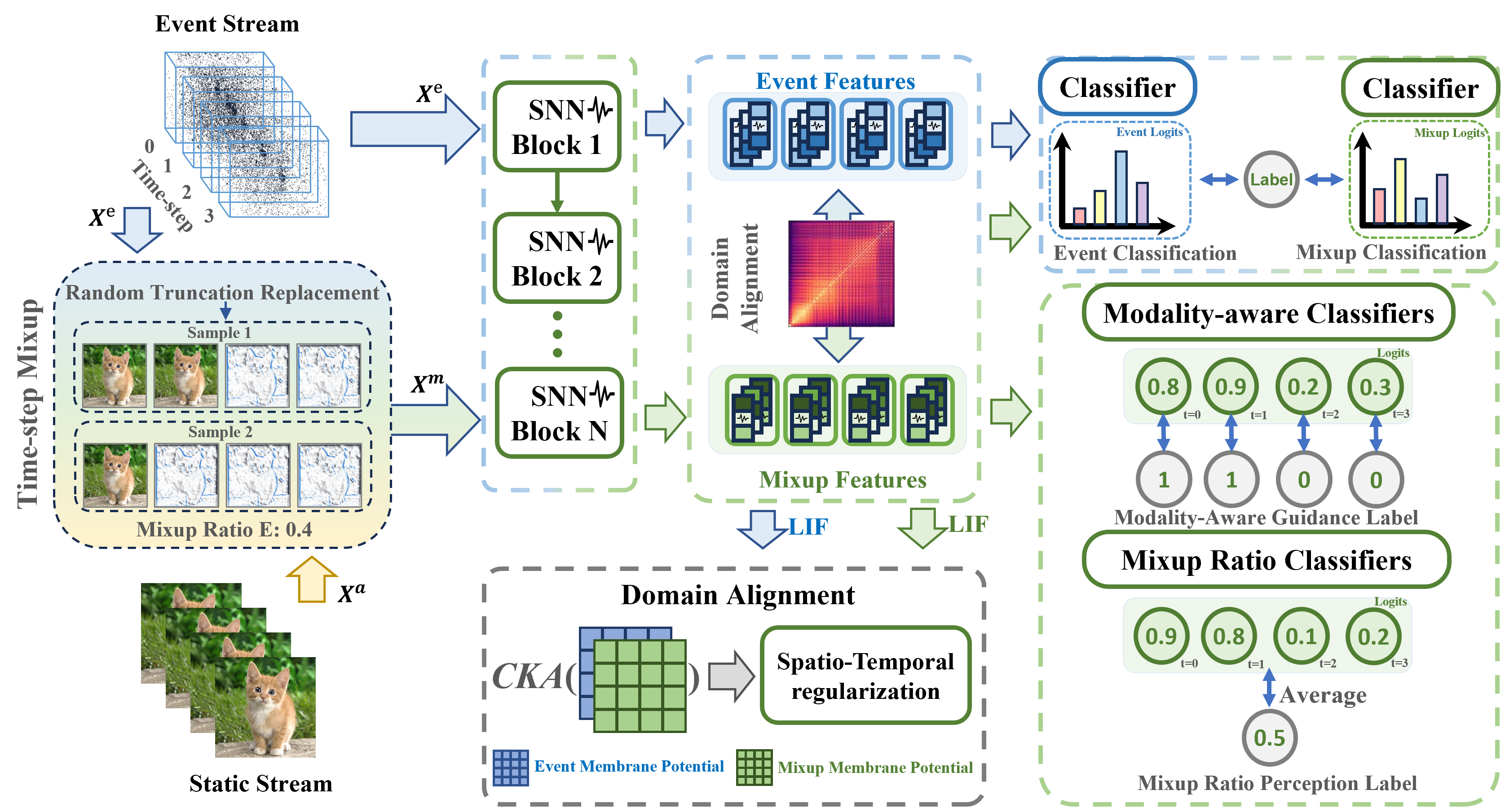} 
\vspace{-7mm}
\caption{The overview of our proposed Time-step Mixup Knowledge Transfer (TMKT) framework. TMKT employs a Time-step Mixup (TSM) strategy and introduces two auxiliary tasks: a modality-aware guidance label and a mixup ratio label to enhance the supervision of temporal knowledge transfer. 
Both the event stream and the Time-step Mixup stream are fed into the network simultaneously, sharing all weights except for the final layer. Membrane potentials from the penultimate layer are used for domain alignment.}
\label{main_fig}
\vspace{-3mm}
\end{figure*}

As shown in Fig.~\ref{main_fig}, our TMKT model adopts a two-stream input paradigm, where paired static and event streams $\mathbf{X}^{\mathrm{a}}$ and $\mathbf{X}^{\mathrm{e}}$ of the same category are provided as input. 
These sequences are first processed by the TSM module to construct time-specific mixed ones $\mathbf{X}^{\mathrm{m}}$, where the modality components are interleaved across time-step.
The resulting sequence $\mathbf{X}^{\mathrm{m}}$ is then forwarded to a SNN-based backbone for feature extraction. 
During this stage, we introduce a Regularized Domain Alignment loss $\mathcal{L}_{\mathrm{RDA}}$ to align the feature distributions of mixed and event modalities within the mixed representation space, mitigating cross-domain discrepancies at the feature level.

To cooperate with the time-step mixed data for effective knowledge transfer, we introduce two novel modality-aware objectives at local and global levels respectively. At each time-step, a MAG loss $\mathcal{L}_{\mathrm{MAG}}$ is crafted to encourage the model to distinguish the dominant modality, promoting temporal consistency across streams. Meanwhile, another MRP loss $\mathcal{L}_{\mathrm{MRP}}$ is proposed to offer global supervision by estimating
the underlying mixing ratio applied by TSM.

\subsection{Time-step Mixup (TSM) Strategy}
Before we dive into the details of Time-step Mixup, let's recall the \textbf{general pipeline of SNNs}. The static or event input are usually in the form frame sequences, denoted as $\mathbf{X}^{\mathrm{a}}=\left\{ \mathbf{x}^{\mathrm{a}}_{1}, \mathbf{x}^{\mathrm{a}}_{2}, \ldots, \mathbf{x}^{\mathrm{a}}_{T} \right\}$ and $\mathbf{X}^{\mathrm{e}}=\left\{ \mathbf{x}^{\mathrm{e}}_{1}, \mathbf{x}^{\mathrm{e}}_{2}, \ldots, \mathbf{x}^{\mathrm{e}}_{T} \right\}$ respectively. \( T \) is the total number of discrete Time-step.
Passing the full sequence \(\mathbf{X}^{\mathrm e}\) through an \(N\)-block SNN encoder $Enc^{(N)}$ yields a \(T\)-step feature sequence:
\begin{equation}
\mathbf{F}^{\mathrm e}\!=\!Enc^{(N)}(\mathbf{X}^{\mathrm e})\!=\!\{\mathbf{F}^{\mathrm e}_t\}_{t=1}^{T}, 
~\mathbf{F}^{\mathrm e}_t\!=\!\big[Enc^{(N)}(\mathbf{X}^{\mathrm e})\big]_t .
\label{eq5}
\end{equation}
A classifier \( \mathcal{F}(\cdot) \) is followed to obtain temporal logits as,
\begin{equation}
\mathbf{y}^{\mathrm{e}}_t = \mathcal{F}(\mathbf{F}^{\mathrm{e}}_{t}). \quad t \in [1, T]
\end{equation}
The temporal efficient training (TET) \cite{tet} loss can be used to integrate Cross-Entropy loss \( CE() \) at each Time-step as:
\begin{equation}
\mathcal{L}_{\mathrm{TET}} = \frac{1}{T} \sum_{t=1}^{T} CE(\mathbf{y}^\mathrm{e}_t, \mathbf{y}),
\label{TET}
\end{equation}
where \( \mathbf{y} \) is the ground truth label.

\vspace{1mm}
\noindent\textbf{Probabilistic Time-step Mixup (TSM).} From above equations, it can be seen that SNNs are naturally trained with sequential data containing multiple Time-step. Thus, our TSM method starts from randomly replacing individual static frames with event ones at different Time-step. 
Noting that, to avoid unstable generalization caused by frequent modality switching across Time-step, we adopt a truncation replacement strategy.
Given an expected Mixup ratio $r_m$, \textit{i.e.,} the portion of replaced samples, we have a corresponding replacement probability $p$ at every time-step, which will be calculated later.
For each static sample, starting from the first time-step $t=1$, we sequentially sample a uniform random variable $u_t \sim \mathcal{U}(0,1)$ at each time-step $t$. If $u_t < p$, we trigger replacement at this frame and substitute all subsequent frames with event-domain data. Formally, the replacement point $t^*$ is determined as,
\begin{equation}
t^* = \min \left\{ t \in \{1, 2, \ldots, T\} \,\middle|\, u_t < p \right\}
 \end{equation}
where it follows a truncated geometric distribution. Given the time-step length $T$, the probability of having $t^*=1,2,\ldots,T$ is calculated as,
\begin{equation}
\mathrm{Pr}(X=t^* \mid X \leq T) = \frac{(1-p)^{t^*-1}p}{1-(1-p)^{T}}
\end{equation}
Then the expectation of the replaced frame number can be computed and it shall align with the Mixup ratio $r_m$ as,
\begin{equation}
\begin{aligned}
\mathbb{E}[X] = \sum_{t=1}^{T}(T+1-t)\cdot\frac{(1-p)^{t-1}p}{1-(1-p)^{T}} = T \cdot {r}_{m}.
\end{aligned}
\label{eq:expect}
\end{equation}
Given the values of $T$ and ${r}_{m}$, the replacement probability $p$ can be obtain by solving Eq. \ref{eq:expect}. Unfortunately, it has no closed-form solution, thus we approximate $p$ using numerical methods\cite{scipy}.
The final mixed input sequence $\mathbf{X}^{\mathrm{m}}=\left\{ \mathbf{x}^{\mathrm{m}}_{1}, \mathbf{x}^{\mathrm{m}}_{2}, \ldots, \mathbf{x}^{\mathrm{m}}_{T} \right\}$ is then constructed as,
 \begin{equation}
 \mathbf{x}_t^{\mathrm{m}} =
 \begin{cases}
 \mathbf{x}_t^\mathrm{a}, & \text{if } \ t < t^* \\
 \mathbf{x}_t^\mathrm{e}, & \text{if } \ t \geq t^*
 \end{cases}.
 \label{eq:mixup}
 \end{equation}
If no replacement occurs, $t^* = T+1$, \textit{i.e.,} the sequence is identical to the original static input.

\subsection{Theoretical Analysis of Time-step Mixup}
We theoretically analyze why such \textbf{Time-step Mixup (TSM)} strategy is better than conventional \textbf{Batch Mixing (BM)}~\cite{ekt}.
We first formalize mild independence assumptions in a unified T-frame setting to compare TSM and BM.

\begin{assumption}[Per–frame gradients]\label{asm:batch}
For each sample $i\!\in\!\{1,\dots,B\}$, there are $T\!=\!n_a\!+\!n_e$ frames. We define $\alpha\!:=\!n_e/T\!\in\![0,1]$
For TSM, the first $n_a$ frames come from the \emph{static} distribution and the last $n_e$ frames from the \emph{event} distribution.
Let $g_a^{i,t},g_e^{i,t}$ denote per–frame gradients for the static and event frame repsectively.
We first assume: \textbf{(i)} samples are i.i.d.\ across $i$; and \textbf{(ii)} we allow general second–order temporal structure within a sample, where gradients for static and event frames are i.d. across time (not necessarily independent), and the temporal covariances are \emph{time–index invariant} (do not depend on the specific $t,s$).

Then, we denote the mean, variance and covariance as:
\begin{align}
\!\mathbb{E}[g_a^{i,t}]\!=\!\mu_a,\mathrm{Var}(g_a^{i,t})\!=\!\Sigma_a,\mathbb{E}[g_e^{i,t}]\!=\!\mu_e,\mathrm{Var}(g_e^{i,t})\!=\!\Sigma_e,\nonumber
\end{align}
\begin{equation}
\begin{split}
&\text{For}~t\neq s:R_a(t,s):=\mathrm{Cov}\big(g_a^{i,t},\,g_a^{i,s}\big),~R_e(t,s):=\\
&\mathrm{Cov}\big(g_e^{i,t},\,g_e^{i,s}\big),~R_{ae}(t,s):=\mathrm{Cov}\big(g_a^{i,t},\,g_e^{i,s}\big),\nonumber
\end{split}
\end{equation}
where $R_a(t,s),R_e(t,s),R_{ae}(t,s)$ are the
\emph{temporal covariance} within the static segment, within the event segment, and the
\emph{cross–segment} covariance, respectively. Moreover, these covariances are \emph{time–index invariant}, i.e., $R_a(t,s)\equiv R_a$, $R_e(t,s)\equiv R_e$, and $R_{ae}(t,s)\equiv R_{ae}$.
We then formulate the per–sample averages for TSM and BM as:
\[
\!Z_i^{\mathrm{TSM}}\!\!=\!\frac{1}{T}\!\!\left(\!\sum_{t=1}^{n_a} g_a^{i,t}\!+\!\!\!\sum_{t=n_a\!+\!1}^{T}\!\! g_e^{i,t}\!\!\right)\!\!,~Z_i^{\mathrm{BM}}\!\!=\!\!
\begin{cases}
\!\frac{1}{T}\!\!\sum_{t\!=\!1}^{T}\! g_a^{i,t}, \!\!\!\!&C_i\!=\!a,\\[4pt]
\!\frac{1}{T}\!\!\sum_{t\!=\!1}^{T} \!g_e^{i,t}, \!\!\!\!&C_i\!=\!e,
\end{cases}
\]
where $C_i\in\{a,e\}$ is a sample–level assignment for BM with $\Pr[C_i=e]=\alpha$. We also assume: \textbf{(iii)} $C_i$ is independent of $\{g_\bullet^{i,t}\}$.
The batch estimators of gradients are: \[G_{\mathrm{TSM}}:=\tfrac{1}{B}\sum_{i=1}^B Z_i^{\mathrm{TSM}},~G_{\mathrm{BM}}:=\tfrac{1}{B}\sum_{i=1}^B Z_i^{\mathrm{BM}}.\]
\end{assumption}

These assumptions are practical and allow us to derive the relationship between the two estimators, $G_{\mathrm{TSM}}$ and $G_{\mathrm{BM}}$, formalized in the next theorem (proof in Appendix~\ref{proof}).

\begin{theorem}[Mean and covariance]\label{thm:tsm-bm}
By Assumption~\ref{asm:batch}, the two estimators share the same expectation:
\begin{equation}
\begin{split}
\mathbb{E}[G_{\mathrm{TSM}}]=\mathbb{E}[G_{\mathrm{BM}}]=(1-\alpha)\,\mu_a+\alpha\,\mu_e.\nonumber
\end{split}
\end{equation}
Their covariance matrices are
\begin{equation}
\begin{split}
&\mathrm{Cov}(G_{\mathrm{TSM}})
\!\!=\!\!\frac{1}{B\,T}\!\Big[
(1\!-\!\alpha)\Sigma_a\!+\!\alpha \Sigma_e
\!+\!(1\!-\!\alpha)((1\!-\!\alpha)T\!\!-\!\!1)R_a
\\
&+\;\alpha\big(\alpha T\!-\!1\big)R_e
+\;(1-\alpha)\big(\alpha T\big)\big(R_{ae}+R_{ae}^{\top}\big)
\Big]\!,\nonumber
\end{split}
\end{equation}
\begin{equation}
\begin{split}
&\mathrm{Cov}(G_{\mathrm{BM}})
=\frac{1}{B\,T}\!\Big[
(1-\alpha)\!\left(\Sigma_a+(T\!-\!1)R_a\right)
\\
&+\!\alpha\!\left(\Sigma_e\!+\!(T\!-\!1)R_e\right)
\Big]
\!+\!\frac{\alpha(1\!-\!\alpha)}{B}(\mu_e\!-\!\mu_a)(\mu_e\!-\!\mu_a)^\top.\nonumber
\end{split}
\end{equation}
Consequently, their difference can be written as
\begin{equation}
\begin{split}
&\mathrm{Cov}(G_{\mathrm{BM}})-\mathrm{Cov}(G_{\mathrm{TSM}})
\\
&=\frac{\alpha(1\!-\!\alpha)}{B}\Big[
(\mu_e\!-\!\mu_a)(\mu_e\!-\!\mu_a)^\top
\!+\! R_a \!+\! R_e \!-\! R_{ae}\!-\!R_{ae}^{\top}
\Big].\nonumber
\end{split}
\end{equation}
\end{theorem}

\begin{remark}\label{remark1}
By Theorem~\ref{thm:tsm-bm}, we have $\mathrm{Cov}(G_{\mathrm{TSM}})\preceq \mathrm{Cov}(G_{\mathrm{BM}})$ without additional assumptions: for any unit vector $v$, $v^\top(\mathrm{Cov}(G_{\mathrm{BM}})-\mathrm{Cov}(G_{\mathrm{TSM}}))v=\tfrac{\alpha(1-\alpha)}{B}\big((v^\top\Delta\mu)^2+v^\top R_\Sigma v\big)\ge 0$, where $\Delta\mu:=\mu_e-\mu_a$ and $R_\Sigma:=R_a+R_e-R_{ae}-R_{ae}^\top\succeq0$ is shown to be PSD by a block-covariance argument (proof in the Appendix~\ref{proof}). Equivalently, the overall noise energy decreases since $\mathrm{tr}(\mathrm{Cov}(G_{\mathrm{BM}})-\mathrm{Cov}(G_{\mathrm{TSM}}))=\tfrac{\alpha(1-\alpha)}{B}\big(\|\Delta\mu\|_2^2+\mathrm{tr}(R_\Sigma)\big)\ge 0$. In practice, larger domain gaps (larger $\|\Delta\mu\|_2$) and weaker cross-segment coupling (smaller $R_{ae}$) can amplify the reduction, but they are not required for the inequality to hold.
\end{remark}

Consequently, from Theorem~\ref{thm:tsm-bm} and Remark~\ref{remark1}, we see that \textbf{TSM yields smaller covariance, lower gradient variance, and thus smoother, more stable optimization}, enabling steadier step sizes and faster convergence. And the gain is largest when the static–event means differ.

\subsection{Domain Alignment}
After obtaining the mixed data $\mathbf{X}^\textrm{m}$, we feed it together with the target-domain (event) input $\mathbf{X}^e$ into a parameter-shared SNN to obtain mixed and event features.
When the membrane potential is below the threshold (Eq.~\ref{eq2}), it is retained without triggering a spike.
The membrane potentials in the last few layers therefore encode rich, high-precision features suitable for feature space alignment.
We thus collect the membrane potentials of the penultimate layer of the SNN feature extractor $Enc^{(N)}(\cdot)$ (Eq.~\ref{eq5}) as a robust common representation space.

We then align this common space with the event feature space to enable effective knowledge transfer.
As illustrated in Fig.~\ref{main_fig}, we employ CKA \cite{cka} to measure representation similarity between the two domains within the SNN (details in Appendix~\ref{Details of CKA}).
At time step $t$, the potentials for the $i$-th mixed sample and the $j$-th event sample are denoted by $\mathbf{V}^\mathrm{m}_{i,t}$ and $\mathbf{V}^\mathrm{e}_{j,t}$, respectively.
The domain alignment loss $\mathcal{L}_{\mathrm{DA}}$ is defined as
\begin{equation}
\mathcal{L}_{\mathrm{DA}}=\frac{1}{T}\sum_{t=1}^T\Bigg(1-
\mathop{CKA}_{\substack{\mathbf{y}_i=\mathbf{y}_j\\ \mathbf{y}\in\mathcal{\mathbf{Y}}}}
\big(\mathbf{V}^\mathrm{m}_{i,t},\mathbf{V}^\mathrm{e}_{j,t}\big)\Bigg),
\label{eqi:domain alignment_1}
\end{equation}
where $\mathbf{y}$ is the class label drawn from the class set $\mathcal{\mathbf{Y}}$, and $\{\mathbf{y}_i=\mathbf{y}_j\}$ denotes matched pairs of mixed and event data.

To avoid overfitting to specific time steps, we further introduce spatio-temporal regularization by assigning learnable weights $\theta_{t}$ to each time step $t$ and using the classification loss ${\mathcal{L}}_{CLS_e} = \mathcal{L}_\text{TET}$ on the event input $\mathbf{X}^{\mathrm{e}}$ as a regularizer.
The regularized domain alignment loss $\mathcal{L}_{\mathrm{RDA}}$ (with $\sigma$ the sigmoid function) is
\begin{equation}
\begin{aligned}
\mathcal{L}_{\mathrm{RDA}}
= \sigma(\theta_{t}) \cdot \mathcal{L}_{\mathrm{DA}}
+ \big(1 - \sigma(\theta_{t})\big)\cdot \frac{1}{T}\sum_{t=1}^{T}\mathcal{L}_{\mathrm{CLS_e}} .
\end{aligned}
\label{equ:domain_alignment}
\end{equation}

\subsection{Auxiliary Supervision from Time-step Mixup}
\label{sec:tsm-aux}
TSM deterministically constructs a mixed sequence
$\mathbf X^{\mathrm m}=\{\mathbf x^{\mathrm m}_t\}_{t=1}^{T}$
with a known transition index $t^\star$ (static $\!\to\!$ event).
This procedure regularizes features but also \emph{creates} two kinds of uncertainty:
(i) \emph{local} ambiguity at each time step about which modality generated the frame, and
(ii) \emph{global} uncertainty about how much of each modality is present across the sequence.
TSM, however, \emph{reveals} both signals by construction.
As shown in Fig.~\ref{main_fig}, we therefore convert them into two lightweight auxiliary tasks that promote mixing-schedule-aware temporal representations.

Let $\mathbf F_t\in\mathbb R^{d}$ denote the SNN feature at time $t$ (\textit{e.g.,} a penultimate membrane/state vector).
We attach two small heads (classifiers):
$g_{\mathrm s}:\mathbb R^{d}\!\to\![0,1]$ for per-frame source prediction and
$g_{\mathrm m}:\mathbb R^{d}\!\to\![0,1]$ for sequence-level ratio estimation.
Both heads are $1$ layer MLPs with sigmoid output.

\vspace{1mm}
\noindent\textbf{Frame-wise Modality-Aware Guidance (MAG).}
To resolve local ambiguity, we supervise the source of each frame using the ground-truth modality revealed by TSM:
\begin{equation}
y^\mathrm{s}_t =
\begin{cases} 
0, & \text{if } \mathbf{x}_t^\mathrm{m} \in \mathbf{X}^\mathrm{e} \\ 
1, & \text{if } \mathbf{x}_t^\mathrm{m} \in \mathbf{X}^\mathrm{a}
\end{cases}.
\label{eq:modality label}
\end{equation}
Given the prediction $\hat z^{\mathrm s}_t=g_{\mathrm s}(\mathbf F_t)$, the MAG loss is
\begin{equation}
\label{eq:mag}
\mathcal L_{\mathrm{MAG}}
=\frac{1}{T}\sum_{t=1}^{T}\mathrm{CE}\!\left(\hat z^{\mathrm s}_t,\,y^{\mathrm s}_t\right).
\end{equation}
MAG supplies per-frame cues and encourages modality-consistent statistics, serving as localized supervision.

\vspace{1mm}
\noindent\textbf{Mixup Ratio Perception (MRP).}
Complementing the local signal, we guide the network to encode the \emph{global} mixing structure.
Let $K=t^\star-1$ be the number of static frames and define the scalar target
$y_{\mathrm m}=K/T\in[0,1]$.
We aggregate per-frame predictions to estimate the sequence-level ratio:
\begin{equation}
\label{eq:mrp}
\hat z_{\mathrm m}=\frac{1}{T}\sum_{t=1}^{T} g_{\mathrm m}(\mathbf F_t),
\quad
\mathcal L_{\mathrm{MRP}}=\mathrm{MSE}\!\left(\hat z_{\mathrm m},\,y_{\mathrm m}\right).
\end{equation}
Complementing the local signal from MAG, MRP encourages temporal features to reflect the global mixing proportion, anchoring how much and how soon one modality dominates the sequence. Together they leverage TSM’s known stability benefits and, as shown in our ablations, are associated with more consistent optimization and accuracy gains.

\subsection{Loss Function}
Overall, the total training loss is defined as,
\begin{equation}
\mathcal{L}_{\mathrm{total}} = \mathcal{L}_{\mathrm{CLS_m}} +\lambda \cdot \mathcal{L}_{\mathrm{RDA}} + \mathcal{L}_{\mathrm{MAG}} + \mathcal{L}_{\mathrm{MRP}},
\label{eq:total loss}
\end{equation}
where $\mathcal{L}_{\mathrm{CLS_m}}$ denotes the classification loss on mixed input sequence $\mathbf{X}^{\mathrm{m}}$, and $\lambda$ is the weighting coefficient for $\mathcal{L}_{\mathrm{RDA}}$.
\begin{table*}[t]
\centering
\caption{Comparison between our TMKT and existing works. The results are mean and standard deviation after taking three different seeds (values in parentheses denote the best result over three runs). Bold and underline indicate the best and second-best results, respectively.}
\vspace{-3mm}
\setlength{\tabcolsep}{8.5pt}
\begin{tabular}{cccccc}
\toprule
\textbf{Dataset}                & \textbf{Category}                    & \textbf{Methods}                              & \textbf{Architecture}             & T                 & \textbf{Accuracy (\%)}                      \\ \midrule
&                                      & NDA \cite{nda}                           & VGGSNN                            & 10                         & 78.2                                   \\
& \multirow{-2}{*}{Data augmentation}  & EventMixer \cite{eventmix}            & ResNet-18                         & 10                         & 79.5                                   \\ \cdashline{2-6} 
&                                      & TET \cite{tet}                          & VGGSNN                            & 10                         & 79.27                                  \\
&                                      & TCJA-TET \cite{tcja}                      & CombinedSNN                       & 14                         & 82.5                                   \\
&                                      & TKS \cite{tks}                    & VGGSNN                            & 10                         & 84.1                                   \\
& \multirow{-4}{*}{Efficient training} & ETC \cite{etc}                          & VGGSNN                            & 10                         & 85.53                                  \\ \cdashline{2-6} 
&                                      & R2ETL with TCKA \cite{stl}              & VGGSNN                            & 10                         & 82.70                                  \\
&                                      & Knowledge-Transfer \cite{ekt}            & VGGSNN                            & 10                         & $93.18\pm0.38$ (93.45)                                  \\
&                                      & CKD \cite{ckd}                                      & VGGSNN                            & 10                         & $96.71\pm0.30$ \underline{(97.13)}                                  \\
\multirow{-10}{*}{N-Caltech101} & \multirow{-4}{*}{Transfer learning}  & \cellcolor[HTML]{EFEFEF}TMKT (Ours) & \cellcolor[HTML]{EFEFEF}VGGSNN    & \cellcolor[HTML]{EFEFEF}10 & \cellcolor[HTML]{EFEFEF}$\textbf{97.70}\pm0.20$ \textbf{(97.93)} \\ \midrule
& Efficient training                   & TET \cite{tet}                          & ResNet-18                         & 6                          & 25.05                                  \\ \cdashline{2-6} 
& Data extension                       & Ev2Vid \cite{ev2vid}                                       & ResNet-18                         & 6                          & \underline{31.20}                                  \\ \cdashline{2-6} 
&                                      & Knowledge-Transfer \cite{ekt}            & ResNet-18                         & 6                          & $30.05\pm0.50$ (30.50)                                \\
\multirow{-4}{*}{CEP-DVS}       & \multirow{-2}{*}{Transfer learning}  & \cellcolor[HTML]{EFEFEF}TMKT (Ours) & \cellcolor[HTML]{EFEFEF}ResNet-18 & \cellcolor[HTML]{EFEFEF}6  & \cellcolor[HTML]{EFEFEF}$\textbf{34.06}\pm0.55$ \textbf{(34.70)}  \\ \midrule
& Efficient training                   & plain \cite{Nomniglot}                         & SCNN                              & 12                         & 60.0                                  \\ \cdashline{2-6} 
&                                      & Knowledge-Transfer \cite{ekt}            & SCNN                              & 12                         & $63.60\pm0.46$ \underline{(64.09)}                                  \\
\multirow{-3}{*}{N-Omniglot}    & \multirow{-2}{*}{Transfer learning}  & \cellcolor[HTML]{EFEFEF}TMKT (Ours) & \cellcolor[HTML]{EFEFEF}SCNN      & \cellcolor[HTML]{EFEFEF}12 & \cellcolor[HTML]{EFEFEF}$\textbf{64.08}\pm0.28$ \textbf{(64.37)}             \\ 
\bottomrule
\end{tabular}
\label{table1}
\vspace{-4mm}
\end{table*}

\section{Experiments}
Our experiments are conducted on several mainstream event-based datasets, including N-Caltech101 \cite{ncal}, and N-Omniglot \cite{Nomniglot}, along with their corresponding RGB-based counterparts. We also conduct experiments on CEP-DVS \cite{cepdvs}, an image-event paired dataset.

\begin{table}[t]
\centering
\caption{Ablation study of components in our TMKT.
}
\vspace{-3mm}
\setlength{\tabcolsep}{3pt}
\begin{tabular}{cccccc}
\toprule
\textbf{Dataset}&\textbf{Network}& \textbf{TSM} & $\mathcal{L}_\mathrm{MAG}$ &{$\mathcal{L}_\mathrm{MRP}$} & \textbf{Accuracy} (\%)\\ 
\cmidrule{1-6}
\multirow{5}{*}{\rotatebox{90}{N-Caltech101}}&{\multirow{5}{*}{\rotatebox{90}{VGGSNN}}}   & - & -  & {-} & 93.18             \\
 && \checkmark & -  & {-} & 97.24 \\
 && \checkmark &\checkmark & {-} & 97.36\\
 && \checkmark    & - &{\checkmark}               & 97.70  \\
&& \checkmark & \checkmark & {\checkmark}  & \textbf{97.93}\\ 
\midrule
\multirow{5}{*}{\rotatebox{90}{CEP-DVS}}&{\multirow{5}{*}{\rotatebox{90}{ResNet-18}}}
& -  & -   & {-} & 30.50             \\
& & \checkmark    & -  & {-}   & 33.00 \\
&& \checkmark    & \checkmark  & {-} & 32.80\\
&& \checkmark    &  -  & {\checkmark}  & 33.55 \\
&& \checkmark & \checkmark &{\checkmark} &\textbf{34.70} \\ 
\bottomrule
\end{tabular}
\label{table2}
\vspace{-2mm}
\end{table}

\begin{table}[t]
\centering
\caption{Ablation study of Time-step Mixup ratio ${r}_{m}$.}
\vspace{-3mm}
\begin{tabular}{cccc}
\toprule
\textbf{Network}                 & \textbf{Dataset}                       & \textbf{Mixup ratio} & \textbf{Accuracy} (\%)       \\ \midrule
\multirow{5}{*}{\rotatebox{90}{VGGSNN}} & \multirow{5}{*}{\rotatebox{90}{N-Caltech101}} & 0.3         & 97.36           \\
                        &                               & 0.4         & \textbf{97.93} \\
                        &                               & 0.5         & 97.59          \\
                        &                               & 0.6         & 97.47            \\
                        &                               & 0.7         & 97.70     \\
\bottomrule
\end{tabular}
\label{table3}
\vspace{-2mm}
\end{table}



\subsection{Experimental Settings}
For a fair comparison, we follow the implementation of our baseline \cite{ekt}. Inputs for N-Caltech101 and CEP-DVS are resized to $48\times48$, and inputs for N-Omniglot are resized to $28\times28$, for both event streams and their RGB counterparts. Model configurations are as follows: VGGSNN\cite{vggsnn} with 10 time steps trained for 300 epochs on N-Caltech101, ResNet-18\cite{resnet} with 6 time steps trained for 200 epochs on CEP-DVS, and SCNN\cite{ekt} with 12 time steps trained for 50 epochs on N-Omniglot. 

Following the baseline configuration, for input encoding, static RGB images are transformed into the HSV color space to reduce the mismatch with event data. Considering the two polarities of events, we replicate the Value channel to match the dual-channel structure and uniformly duplicate the static image across time steps. The Mixup ratio $r_m$ is set to $0.4$, $0.9$, and $0.95$ on N-Caltech101, CEP-DVS, and N-Omniglot, respectively. The coefficient $\lambda$ in Eq.~\ref{eq:total loss} is fixed to $0.5$ for all experiments. All experiments are implemented with the BrainCog \cite{braincog}. Unless otherwise stated, all reported results are the mean $\pm$ standard deviation over three runs with different random seeds.

\subsection{Comparison with the State-of-the-Art}
The experimental results on N-Caltech101, CEP-DVS, and N-Omniglot are summarized in Tab.~\ref{table1} for detailed comparison. The method categorization follows the baseline \cite{ekt}.
On N-Caltech101, our method is compared with several recent state-of-the-art approaches, including three different categories (data augmentation, efficient learning and transfer learning).
Under the VGGSNN architecture, our method achieves a new state-of-the-art accuracy of 97.93\%, demonstrating the superior effectiveness of the proposed TMKT framework. It outperforms our baseline method Knowledge-Transfer \cite{ekt} by a notable margin of 4.75\%.  On CEP-DVS, we also observe a constant performance gain, achieving a 4.2\% improvement over the baseline.

N-Omniglot is a challenging few-shot event-based dataset with limited samples per class and substantial noise/artifacts.
Under the same setup, we reproduce only 60.69\% accuracy for the baseline, whereas TMKT attains 64.37\%, surpassing the baseline-reported result and highlighting its effectiveness in the few-shot setting.




\subsection{Ablation Study}
To verify the effectiveness of our method, extensive ablation studies are conducted comparing to our baseline Knowledge-Transfer (Batch Mixing) \cite{ekt}.

\vspace{1mm}
\noindent\textbf{Time-step Mixup Knowledge Transfer.}
We evaluated our proposed TMKT on both VGGSNN and ResNet-18, with the results summarized in Tab. \ref{table2}. 
Compared to the baseline, applying the Time-step Mixup (TSM) alone yields a significant performance improvement, demonstrating its effectiveness in helping the model smoothly learn from a common representation space across modalities. 
When supervised by the two proposed auxiliary losses, one of which indicates the dominant modality at each timestep while the other encodes the overall mixup ratio of multimodal information for the entire sample, the model consistently achieves improved performance. When combined, these two losses lead to the best overall results.

\vspace{1mm}
\noindent\textbf{Mixup ratio.}
The mixup ratio \( r_{m} \) is a crucial hyperparameter in our framework, as it determines the overall mixing proportion between static data and event data in the Time-step Mixup process.  
We conduct ablation experiments on the N-Caltech101 dataset, as shown in Tab.~\ref{table3}, and find that \( r_{m} = 0.4 \) yields the best performance.  
Notably, all other ablated settings still significantly outperform the baseline Knowledge-Transfer \cite{ekt}, further demonstrating its effectiveness and robustnes.

\begin{table}[t]
\centering
\caption{Ablation study of Replacement-Probability Schedule.}
\vspace{-3mm}
\setlength{\tabcolsep}{4.5pt}
\begin{tabular}{ccc}
\toprule
\textbf{Network}        & \textbf{Schedule} & \textbf{Accuracy}  (\%)\\ \midrule
\multirow{4}{*}{\rotatebox{90}{VGGSNN}} & Fixed Ratio             &  95.86                \\
& Dynamic Ratio (Non-Linear) &  95.05           \\
& Dynamic Ratio (Linear)  &  96.55           \\                        
& Our Probabilistic TSM         & \textbf{97.93}    \\ \bottomrule
\end{tabular}
\label{table4}
\vspace{-2mm}
\end{table}

\begin{table}[t]
\centering
\caption{Ablation of TMH Schedules.}
\vspace{-4mm}
\setlength{\tabcolsep}{2pt}
\begin{tabular}{ccc}
\toprule
\textbf{Network} & \textbf{Schedule} & \textbf{Accuracy} (\%) \\ \midrule
\multirow{4}{*}{\rotatebox{90}{VGGSNN}} 
& TMH-R$\to$D (ours)      & \textbf{97.93} \\
& TMH-D$\to$R             & 97.01 \\
& TMH-Mid(DVS)            & 97.70 \\                       
& TMH-Rand(DVS)           & 97.59 \\                  
\midrule
\end{tabular}
\label{table5}
\vspace{-2mm}
\end{table}


\vspace{1mm}

\noindent\textbf{Mixup strategy.}
We explore Time-step Mixup (TSM) from two perspectives: \emph{how much} to replace (mixup ratio) and \emph{where} to replace (replacement position).
As shown in Tab.~\ref{table4}, we compare three mixup ratio designs: \emph{fixed ratio}, \emph{dynamic ratio} (linear/nonlinear), and our \emph{probabilistic TSM}.
In the fixed ratio case, the switch point $t^{*}$ is set to $\lfloor T \cdot r_{m} \rfloor$.
The dynamic ratio progressively increases the replacement during training as
\begin{align}
r_{m}' &= \left[\frac{b_{i}+e_{c}\,b_{l}}{e_{m}\,b_{l}}\right]^{3}, \quad
r_{m}'' = \frac{e_{c}}{e_{m}},
\end{align}
where $b_i$ is the batch index, $b_l$ the number of batches per epoch, $e_c$ the current epoch, and $e_m$ the total number of epochs.
Our \emph{probabilistic TSM} instead samples $t^{*}$ for each sequence from a truncated geometric distribution whose expectation matches $r_m$, offering diverse boundary coverage while maintaining a smooth RGB$\rightarrow$DVS transition.
Empirically, probabilistic TSM outperforms all other schedules in Tab.~\ref{table4}, indicating that diverse yet unbiased temporal mixup improves robustness and generalization.

Beyond the ratio, we also investigate where the handoff occurs (Tab.~\ref{table5}) using four \emph{Temporal Modality Handoff (TMH)} schedules on VGGSNN: \emph{TMH-R$\rightarrow$D} (Early-RGB / Late-DVS), \emph{TMH-D$\rightarrow$R} (Early-DVS / Late-RGB), \emph{TMH-Mid(DVS)} (Mid-Window DVS), and \emph{TMH-Rand(DVS)} (Random-Step DVS).
\emph{TMH-R$\rightarrow$D} achieves the highest accuracy of \textbf{97.93\%}, as early dense RGB frames stabilize membrane integration and provide class-discriminative cues, while later sparse, motion-sensitive DVS frames preserve the static-to-event causality and mitigate cross-modal drift.

\begin{figure}[t]
\centering
    \begin{subfigure}{0.49\linewidth}
        \centering
        \includegraphics[width=\linewidth]{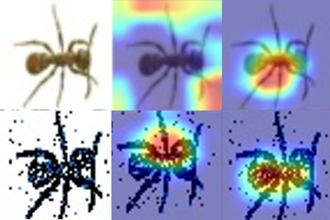}
        \caption{ant}
    \end{subfigure}
    \hfill
    \begin{subfigure}{0.49\linewidth}
        \centering
        \includegraphics[width=\linewidth]{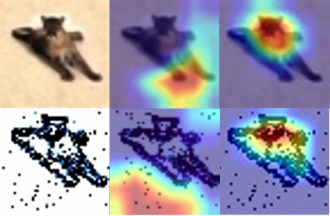}
        \caption{cougar body}
    \end{subfigure}
    \hfill
    \begin{subfigure}{0.49\linewidth}
        \centering
        \includegraphics[width=\linewidth]{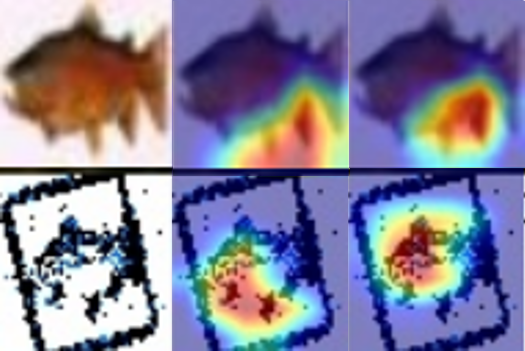}
        \caption{fish}
    \end{subfigure}
    \hfill
    \begin{subfigure}{0.49\linewidth}
        \centering
        \includegraphics[width=\linewidth]{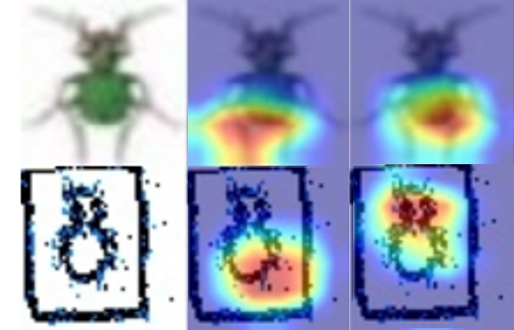}
        \caption{insects}
    \end{subfigure}
    \vspace{-3mm}
\caption{Class Activation Mapping of N-Caltech101 (a)(b), CEP-DVS (c)(d), and their RGB counterparts. For each class, the top row shows static images, and the bottom row presents event data integrated into frames. Within each class, from left to right are: original input, baseline \cite{ekt} result, and our result.}
\label{fig:cam}
\vspace{-2mm}
\end{figure}

\begin{figure}[t]
\centering
    \begin{subfigure}{0.49\linewidth}
        \centering
        \includegraphics[width=\linewidth]{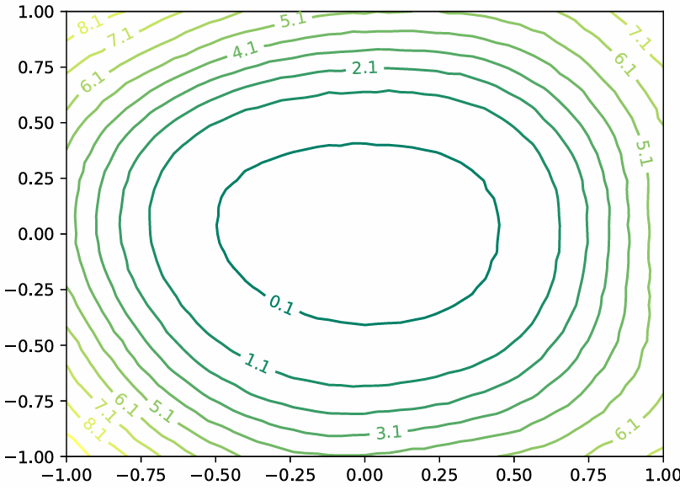}
        \caption{Baseline,N-Caltech101 \cite{ekt}}
    \end{subfigure}
    \hfill
    \begin{subfigure}{0.49\linewidth}
        \centering
        \includegraphics[width=\linewidth]{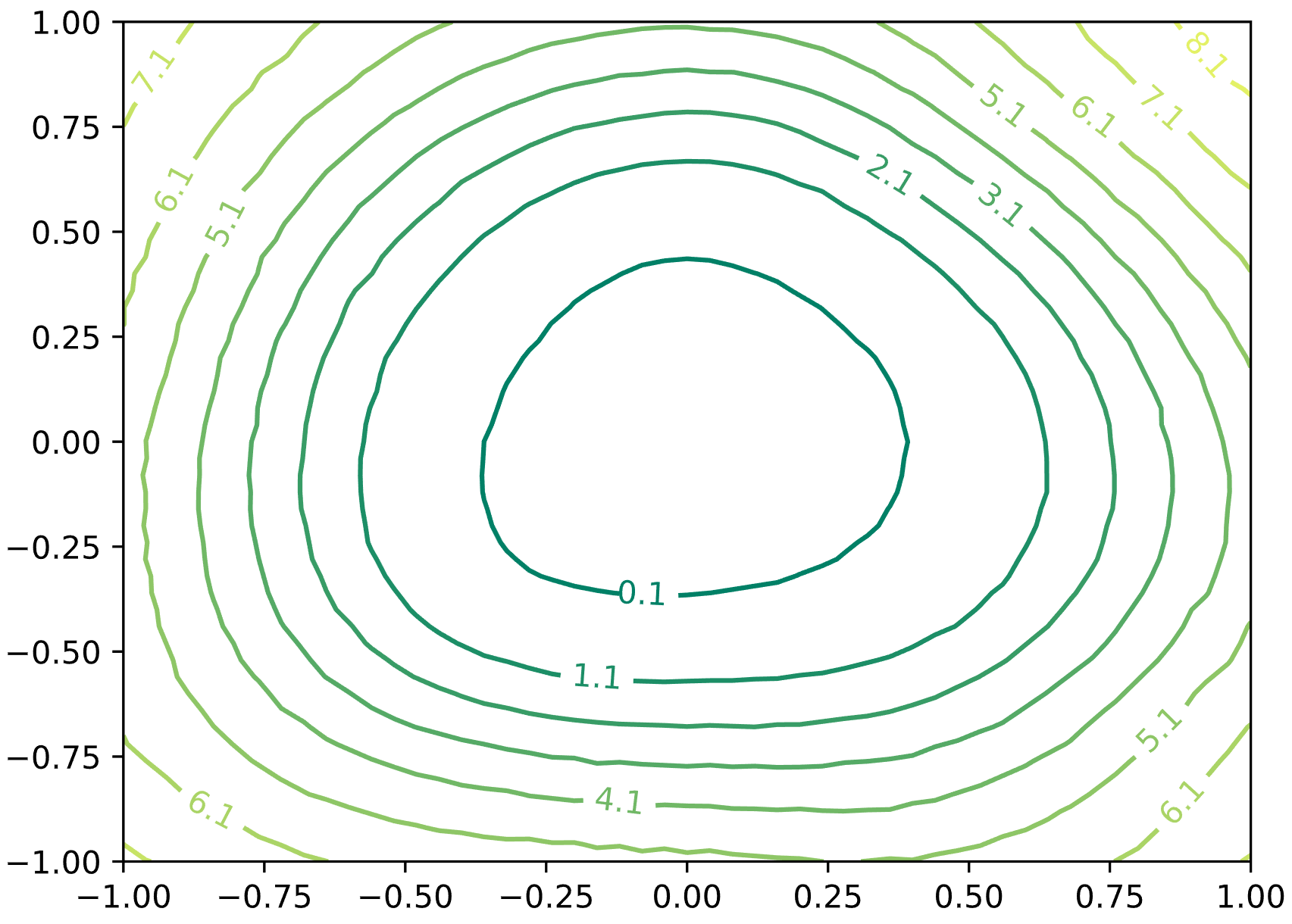}
        \caption{Ours,N-Caltech101}
    \end{subfigure}
    \hfill
    \begin{subfigure}{0.49\linewidth}
        \centering
        \includegraphics[width=\linewidth]{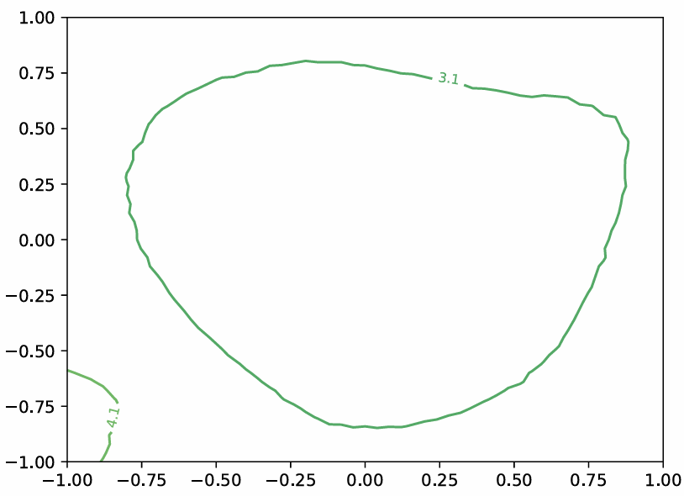}
        \caption{Baseline,CEP-DVS \cite{ekt}}
    \end{subfigure}
    \hfill
    \begin{subfigure}{0.49\linewidth}
        \centering
        \includegraphics[width=\linewidth]{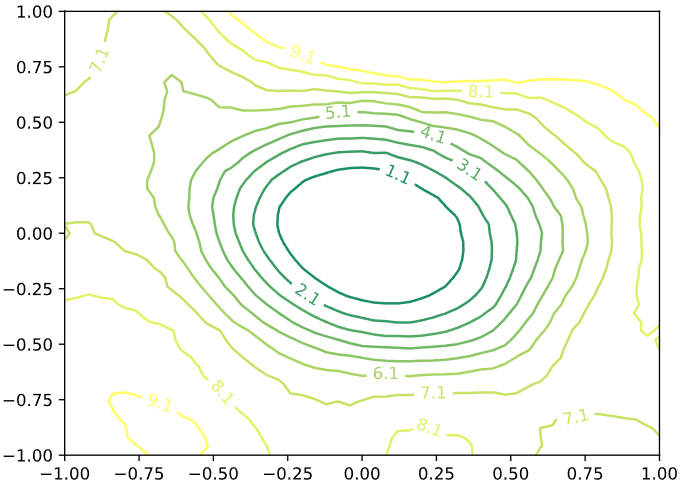}
        \caption{Ours,N-Caltech101}
    \end{subfigure}
    \vspace{-3mm}
\caption{Visualization of the loss landscapes for our method and the baseline Knowledge-Transfer \cite{ekt}.}
\label{fig:loss_landscape}
\vspace{-2mm}
\end{figure}

\subsection{Analysis and Discussion}
\noindent\textbf{Cross-Modal Visual Interpretability.}
To further assess whether our method successfully learns a common representational space across the static and event domains, we adopt grad-cam++ \cite{grad_cam} for visual explanation, highlighting the image regions that contribute most to the final classification decision. As shown in Fig. \ref{fig:cam}, our method consistently focuses on the key semantic regions in both static and event inputs, demonstrating its ability to bridge modality differences and extract shared discriminative features.

\vspace{1mm}
\noindent\textbf{Loss Landscape.}
To investigate whether our method enables the SNN to learn more discriminative features in the event domain, we perform experiments using 2D loss landscape visualization \cite{losslandscape} on the N-Caltech101 and CEP-DVS datasets, comparing our method with the baseline.  
As demonstrated in Fig.~\ref{fig:loss_landscape}, our approach produces a more compact and concentrated loss basin around the minimum, suggesting that the model converges to a sharper and more well-defined solution.  
This indicates that our training strategy facilitates the learning of more discriminative representations within the event domain.  
In contrast, the baseline exhibits a flatter and more irregular surface. It may reflect a less stable convergence behavior and potentially inferior generalization capability.
\section{Conclusion}
In this paper, we proposed a Time-step Mixup Knowledge Transfer (TMKT) framework for spiking neural networks to transfer knowledge from the static domain to the event domain. By mixing static and event sequences at the time-step level and introducing Modality-aware Guidance, Mixup Ratio Perception, and Domain Alignment, TMKT achieves smooth and effective cross-modal knowledge transfer. Experiments on N-Caltech101, CEP-DVS, and other datasets demonstrate superior performance, suggesting TMKT as a new paradigm for SNN-based knowledge transfer. The future work will focus on more different and dedicated mixing strategies over both spatial and temporal domains.

\newpage
{
    \small
    \bibliographystyle{ieeenat_fullname}
    \bibliography{main}

@String(ICML= {Int. Conf. Mach. Learn.})

@String(NIPS= {Adv. Neural Inform. Process. Syst.})

@String(AAAI = {AAAI})

@String(CVPRW= {IEEE Conf. Comput. Vis. Pattern Recog. Worksh.})

@String(NIPS  = {NeurIPS})

@String(CVPRW= {CVPRW})

@inproceedings{rebecq2019events,
  title={Events-to-video: Bringing modern computer vision to event cameras},
  author={Rebecq, Henri and Ranftl, Ren{\'e} and Koltun, Vladlen and Scaramuzza, Davide},
  booktitle={Proceedings of the IEEE/CVF Conference on Computer Vision and Pattern Recognition},
  pages={3857--3866},
  year={2019}
}

@article{zhu2018multivehicle,
  title={The multivehicle stereo event camera dataset: An event camera dataset for 3D perception},
  author={Zhu, Alex Zihao and Thakur, Dinesh and {\"O}zaslan, Tolga and Pfrommer, Bernd and Kumar, Vijay and Daniilidis, Kostas},
  journal={IEEE Robotics and Automation Letters},
  volume={3},
  number={3},
  pages={2032--2039},
  year={2018},
  publisher={IEEE}
}

@article{gehrig2021dsec,
  title={Dsec: A stereo event camera dataset for driving scenarios},
  author={Gehrig, Mathias and Aarents, Willem and Gehrig, Daniel and Scaramuzza, Davide},
  journal={IEEE Robotics and Automation Letters},
  volume={6},
  number={3},
  pages={4947--4954},
  year={2021},
  publisher={IEEE}
}

@inproceedings{gehrig2020video,
  title={Video to events: Recycling video datasets for event cameras},
  author={Gehrig, Daniel and Gehrig, Mathias and Hidalgo-Carri{\'o}, Javier and Scaramuzza, Davide},
  booktitle={Proceedings of the IEEE/CVF Conference on Computer Vision and Pattern Recognition},
  pages={3586--3595},
  year={2020}
}

@article{kudithipudi2025neuromorphic,
  title={Neuromorphic computing at scale},
  author={Kudithipudi, Dhireesha and Schuman, Catherine and Vineyard, Craig M and Pandit, Tej and Merkel, Cory and Kubendran, Rajkumar and Aimone, James B and Orchard, Garrick and Mayr, Christian and Benosman, Ryad and others},
  journal={Nature},
  volume={637},
  number={8047},
  pages={801--812},
  year={2025},
  publisher={Nature Publishing Group UK London}
}

@inproceedings{deng2009imagenet,
  title={Imagenet: A large-scale hierarchical image database},
  author={Deng, Jia and Dong, Wei and Socher, Richard and Li, Li-Jia and Li, Kai and Fei-Fei, Li},
  booktitle={2009 IEEE conference on computer vision and pattern recognition},
  pages={248--255},
  year={2009},
  organization={Ieee}
}

@article{gallego2020event,
  title={Event-based vision: A survey},
  author={Gallego, Guillermo and Delbr{\"u}ck, Tobi and Orchard, Garrick and Bartolozzi, Chiara and Taba, Brian and Censi, Andrea and Leutenegger, Stefan and Davison, Andrew J and Conradt, J{\"o}rg and Daniilidis, Kostas and others},
  journal={IEEE transactions on pattern analysis and machine intelligence},
  volume={44},
  number={1},
  pages={154--180},
  year={2020},
  publisher={IEEE}
}

@article{merolla2014truenorth,
  title   = {A million spiking-neuron integrated circuit with a scalable communication network and interface},
  author  = {Merolla, Paul A. and Arthur, John V. and others},
  journal = {Science},
  volume  = {345},
  number  = {6197},
  pages   = {668--673},
  year    = {2014}
}

@inproceedings{li2021dspike,
  title={Differentiable Spike: Rethinking Gradient-Descent for Training Spiking Neural Networks},
  author={Li, Yuhang and Guo, Yufei and Zhang, Shanghang and Deng, Shikuang and Hai, Yongqing and Gu, Shi},
  booktitle=NIPS,
  year={2021}
}

@inproceedings{wang2023asgl,
  title={Adaptive Smoothing Gradient Learning for Spiking Neural Networks},
  author={Wang, Ziming and Zhang, Wanli and Zhao, Guoqiang and others},
  booktitle=ICML,
  series={PMLR},
  volume={202},
  pages={36879--36903},
  year={2023}
}

@inproceedings{zheng2021tdbn,
  title={Going Deeper with Directly-Trained Larger Spiking Neural Networks},
  author={Zheng, Hanle and Wu, Yujie and Deng, Lei and Hu, Xing and Li, Guoqi},
  booktitle=AAAI,
  year={2021},
  pages={11062--11070}
}

@article{roy,
  title={Towards spike-based machine intelligence with neuromorphic computing},
  author={Roy, Kaushik and Jaiswal, Akhilesh and Panda, Priyadarshini},
  journal={Nature},
  volume={575},
  number={7784},
  pages={607--617},
  year={2019},
  publisher={Nature Publishing Group UK London}
}

@article{davies2018loihi,
  title   = {Loihi: A Neuromorphic Manycore Processor with On-Chip Learning},
  author  = {Davies, Mike and Srinivasa, Narayan and Lin, Tsung-Han and Chinya, Gautham and others},
  journal = {IEEE Micro},
  volume  = {38},
  number  = {1},
  pages   = {82--99},
  year    = {2018},
  doi     = {10.1109/MM.2018.112130359}
}

@article{lenero20113,
  title={A 3.6 $\mu$ s latency asynchronous frame-free event-driven dynamic-vision-sensor},
  author={Le{\~n}ero-Bardallo, Juan Antonio and Serrano-Gotarredona, Teresa and Linares-Barranco, Bernab{\'e}},
  journal={IEEE Journal of Solid-State Circuits},
  volume={46},
  number={6},
  pages={1443--1455},
  year={2011},
  publisher={IEEE}
}

@article{davis,
  title={A 240$\times$ 180 130 db 3 $\mu$s latency global shutter spatiotemporal vision sensor},
  author={Brandli, Christian and Berner, Raphael and Yang, Minhao and Liu, Shih-Chii and Delbruck, Tobi},
  journal={IEEE Journal of Solid-State Circuits},
  volume={49},
  number={10},
  pages={2333--2341},
  year={2014},
  publisher={IEEE}
}

@inproceedings{celex,
  title={Live demonstration: CeleX-V: A 1M pixel multi-mode event-based sensor},
  author={Chen, Shoushun and Guo, Menghan},
  booktitle={2019 IEEE/CVF Conference on Computer Vision and Pattern Recognition Workshops (CVPRW)},
  pages={1682--1683},
  year={2019},
  organization={IEEE}
}

@article{atis,
  title={A QVGA 143 dB dynamic range frame-free PWM image sensor with lossless pixel-level video compression and time-domain CDS},
  author={Posch, Christoph and Matolin, Daniel and Wohlgenannt, Rainer},
  journal={IEEE Journal of Solid-State Circuits},
  volume={46},
  number={1},
  pages={259--275},
  year={2010},
  publisher={IEEE}
}

@inproceedings{ucfcrimedvs,
  title={UCF-Crime-DVS: A Novel Event-Based Dataset for Video Anomaly Detection with Spiking Neural Networks},
  author={Qian, Yuanbin and Ye, Shuhan and Wang, Chong and Cai, Xiaojie and Qian, Jiangbo and Wu, Jiafei},
  booktitle={Proceedings of the AAAI Conference on Artificial Intelligence},
  volume={39},
  number={6},
  pages={6577--6585},
  year={2025}
}

@article{deng2021,
  title={Learning from images: A distillation learning framework for event cameras},
  author={Deng, Yongjian and Chen, Hao and Chen, Huiying and Li, Youfu},
  journal={IEEE Transactions on Image Processing},
  volume={30},
  pages={4919--4931},
  year={2021},
  publisher={IEEE}
}

@inproceedings{ekt,
  title={An efficient knowledge transfer strategy for spiking neural networks from static to event domain},
  author={He, Xiang and Zhao, Dongcheng and Li, Yang and Shen, Guobin and Kong, Qingqun and Zeng, Yi},
  booktitle={Proceedings of the AAAI Conference on Artificial Intelligence},
  volume={38},
  number={1},
  pages={512--520},
  year={2024}
}

@article{stl,
  title={Spiking transfer learning from rgb image to neuromorphic event stream},
  author={Zhan, Qiugang and Liu, Guisong and Xie, Xiurui and Tao, Ran and Zhang, Malu and Tang, Huajin},
  journal={IEEE Transactions on Image Processing},
  year={2024},
  publisher={IEEE}
}

@inproceedings{mixup,
  title={mixup: Beyond Empirical Risk Minimization},
  author={Zhang, Hongyi and Cisse, Moustapha and Dauphin, Yann N and Lopez-Paz, David},
  booktitle={International Conference on Learning Representations},
  year={2018}
}

@misc{ckd,
      title={Cross Knowledge Distillation between Artificial and Spiking Neural Networks}, 
      author={Shuhan Ye and Yuanbin Qian and Chong Wang and Sunqi Lin and Jiazhen Xu and Jiangbo Qian and Yuqi Li},
      year={2025},
      eprint={2507.09269},
      archivePrefix={arXiv},
      primaryClass={cs.CV},
      url={https://arxiv.org/abs/2507.09269}, 
}

@inproceedings{nda,
  title={Neuromorphic data augmentation for training spiking neural networks},
  author={Li, Yuhang and Kim, Youngeun and Park, Hyoungseob and Geller, Tamar and Panda, Priyadarshini},
  booktitle={European Conference on Computer Vision},
  pages={631--649},
  year={2022},
  organization={Springer}
}

@article{eventmix,
  title={Eventmix: An efficient data augmentation strategy for event-based learning},
  author={Shen, Guobin and Zhao, Dongcheng and Zeng, Yi},
  journal={Information Sciences},
  volume={644},
  pages={119170},
  year={2023},
  publisher={Elsevier}
}

@article{tet,
  title={Temporal efficient training of spiking neural network via gradient re-weighting},
  author={Deng, Shikuang and Li, Yuhang and Zhang, Shanghang and Gu, Shi},
  journal={arXiv preprint arXiv:2202.11946},
  year={2022}
}

@article{tcja,
  title={Tcja-snn: Temporal-channel joint attention for spiking neural networks},
  author={Zhu, Rui-Jie and Zhang, Malu and Zhao, Qihang and Deng, Haoyu and Duan, Yule and Deng, Liang-Jian},
  journal={IEEE Transactions on Neural Networks and Learning Systems},
  volume={36},
  number={3},
  pages={5112--5125},
  year={2024},
  publisher={IEEE}
}

@article{tks,
  author       = {Yiting Dong and
                  Dongcheng Zhao and
                  Yi Zeng},
  title        = {Temporal Knowledge Sharing Enable Spiking Neural Network Learning
                  From Past and Future},
  journal      = {{IEEE} Trans. Artif. Intell.},
  volume       = {5},
  number       = {7},
  pages        = {3524--3534},
  year         = {2024},
  url          = {https://doi.org/10.1109/TAI.2024.3374268},
  doi          = {10.1109/TAI.2024.3374268},
  bibsource    = {dblp computer science bibliography, https://dblp.org}
}

@article{etc,
  author       = {Dongcheng Zhao and
                  Guobin Shen and
                  Yiting Dong and
                  Yang Li and
                  Yi Zeng},
  title        = {Improving stability and performance of spiking neural networks through
                  enhancing temporal consistency},
  journal      = {Pattern Recognit.},
  volume       = {159},
  pages        = {111094},
  year         = {2025},
  url          = {https://doi.org/10.1016/j.patcog.2024.111094},
  doi          = {10.1016/J.PATCOG.2024.111094},
  bibsource    = {dblp computer science bibliography, https://dblp.org}
}

@article{Nomniglot,
  title={N-omniglot, a large-scale neuromorphic dataset for spatio-temporal sparse few-shot learning},
  author={Li, Yang and Dong, Yiting and Zhao, Dongcheng and Zeng, Yi},
  journal={Scientific Data},
  volume={9},
  number={1},
  pages={746},
  year={2022},
  publisher={Nature Publishing Group UK London}
}

@article{cepdvs,
  title={Learning from images: A distillation learning framework for event cameras},
  author={Deng, Yongjian and Chen, Hao and Chen, Huiying and Li, Youfu},
  journal={IEEE Transactions on Image Processing},
  volume={30},
  pages={4919--4931},
  year={2021},
  publisher={IEEE}
}

@article{ev2vid,
  title={High speed and high dynamic range video with an event camera},
  author={Rebecq, Henri and Ranftl, Ren{\'e} and Koltun, Vladlen and Scaramuzza, Davide},
  journal={IEEE transactions on pattern analysis and machine intelligence},
  volume={43},
  number={6},
  pages={1964--1980},
  year={2019},
  publisher={IEEE}
}

@article{ncal,
  title={Converting static image datasets to spiking neuromorphic datasets using saccades},
  author={Orchard, Garrick and Jayawant, Ajinkya and Cohen, Gregory K and Thakor, Nitish},
  journal={Frontiers in neuroscience},
  volume={9},
  pages={437},
  year={2015},
  publisher={Frontiers Media SA}
}

@inproceedings{cka,
  title={Similarity of neural network representations revisited},
  author={Kornblith, Simon and Norouzi, Mohammad and Lee, Honglak and Hinton, Geoffrey},
  booktitle={International conference on machine learning},
  pages={3519--3529},
  year={2019},
  organization={PMlR}
}

@article{braincog,
  title={Braincog: A spiking neural network based, brain-inspired cognitive intelligence engine for brain-inspired ai and brain simulation},
  author={Zeng, Yi and Zhao, Dongcheng and Zhao, Feifei and Shen, Guobin and Dong, Yiting and Lu, Enmeng and Zhang, Qian and Sun, Yinqian and Liang, Qian and Zhao, Yuxuan and others},
  journal={Patterns},
  volume={4},
  number={8},
  year={2023},
  publisher={Elsevier}
}

@book{LIF,
  title={Neuronal dynamics: From single neurons to networks and models of cognition},
  author={Gerstner, Wulfram and Kistler, Werner M and Naud, Richard and Paninski, Liam},
  year={2014},
  publisher={Cambridge University Press}
}

@article{wang2019domain,
  title={Domain adaptation with neural embedding matching},
  author={Wang, Zengmao and Du, Bo and Guo, Yuhong},
  journal={IEEE transactions on neural networks and learning systems},
  volume={31},
  number={7},
  pages={2387--2397},
  year={2019},
  publisher={IEEE}
}

@inproceedings{guo2019mixup,
  title={Mixup as locally linear out-of-manifold regularization},
  author={Guo, Hongyu and Mao, Yongyi and Zhang, Richong},
  booktitle={Proceedings of the AAAI conference on artificial intelligence},
  volume={33},
  number={01},
  pages={3714--3722},
  year={2019}
}

@article{hu2021neural,
  title={Neural dubber: Dubbing for videos according to scripts},
  author={Hu, Chenxu and Tian, Qiao and Li, Tingle and Yuping, Wang and Wang, Yuxuan and Zhao, Hang},
  journal={Advances in neural information processing systems},
  volume={34},
  pages={16582--16595},
  year={2021}
}

@inproceedings{wang2022vlmixer,
  title={Vlmixer: Unpaired vision-language pre-training via cross-modal cutmix},
  author={Wang, Teng and Jiang, Wenhao and Lu, Zhichao and Zheng, Feng and Cheng, Ran and Yin, Chengguo and Luo, Ping},
  booktitle={International Conference on Machine Learning},
  pages={22680--22690},
  year={2022},
  organization={PMLR}
}

@inproceedings{losslandscape,
  author       = {Hao Li and
                  Zheng Xu and
                  Gavin Taylor and
                  Christoph Studer and
                  Tom Goldstein},
  editor       = {Samy Bengio and
                  Hanna M. Wallach and
                  Hugo Larochelle and
                  Kristen Grauman and
                  Nicol{\`{o}} Cesa{-}Bianchi and
                  Roman Garnett},
  title        = {Visualizing the Loss Landscape of Neural Nets},
  booktitle    = {Advances in Neural Information Processing Systems 31: Annual Conference
                  on Neural Information Processing Systems 2018, NeurIPS 2018, December
                  3-8, 2018, Montr{\'{e}}al, Canada},
  pages        = {6391--6401},
  year         = {2018},
  url          = {https://proceedings.neurips.cc/paper/2018/hash/a41b3bb3e6b050b6c9067c67f663b915-Abstract.html},
  bibsource    = {dblp computer science bibliography, https://dblp.org}
}

@inproceedings{grad_cam,
  author       = {Aditya Chattopadhyay and
                  Anirban Sarkar and
                  Prantik Howlader and
                  Vineeth N. Balasubramanian},
  title        = {Grad-CAM++: Generalized Gradient-Based Visual Explanations for Deep
                  Convolutional Networks},
  booktitle    = {2018 {IEEE} Winter Conference on Applications of Computer Vision,
                  {WACV} 2018, Lake Tahoe, NV, USA, March 12-15, 2018},
  pages        = {839--847},
  publisher    = {{IEEE} Computer Society},
  year         = {2018},
  url          = {https://doi.org/10.1109/WACV.2018.00097},
  doi          = {10.1109/WACV.2018.00097},
  bibsource    = {dblp computer science bibliography, https://dblp.org}
}

@inproceedings{spikformer,
title={Spikformer: When Spiking Neural Network Meets Transformer },
author={Zhaokun Zhou and Yuesheng Zhu and Chao He and Yaowei Wang and Shuicheng YAN and Yonghong Tian and Li Yuan},
booktitle={The Eleventh International Conference on Learning Representations },
year={2023},
url={https://openreview.net/forum?id=frE4fUwz_h}
}

@article{sdv3,
  title={Scaling spike-driven transformer with efficient spike firing approximation training},
  author={Yao, Man and Qiu, Xuerui and Hu, Tianxiang and Hu, Jiakui and Chou, Yuhong and Tian, Keyu and Liao, Jianxing and Leng, Luziwei and Xu, Bo and Li, Guoqi},
  journal={IEEE Transactions on Pattern Analysis and Machine Intelligence},
  year={2025},
  publisher={IEEE}
}

@inproceedings{cutmix,
  title={Cutmix: Regularization strategy to train strong classifiers with localizable features},
  author={Yun, Sangdoo and Han, Dongyoon and Oh, Seong Joon and Chun, Sanghyuk and Choe, Junsuk and Yoo, Youngjoon},
  booktitle={Proceedings of the IEEE/CVF international conference on computer vision},
  pages={6023--6032},
  year={2019}
}

@book{IF,
  title={Spiking neuron models: Single neurons, populations, plasticity},
  author={Gerstner, Wulfram and Kistler, Werner M},
  year={2002},
  publisher={Cambridge university press}
}

@article{scipy,
  title={SciPy 1.0: fundamental algorithms for scientific computing in Python},
  author={Virtanen, Pauli and Gommers, Ralf and Oliphant, Travis E and Haberland, Matt and Reddy, Tyler and Cournapeau, David and Burovski, Evgeni and Peterson, Pearu and Weckesser, Warren and Bright, Jonathan and others},
  journal={Nature methods},
  volume={17},
  number={3},
  pages={261--272},
  year={2020},
  publisher={Nature Publishing Group US New York}
}

@article{vggsnn,
  title={Going deeper in spiking neural networks: VGG and residual architectures},
  author={Sengupta, Abhronil and Ye, Yuting and Wang, Robert and Liu, Chiao and Roy, Kaushik},
  journal={Frontiers in neuroscience},
  volume={13},
  pages={95},
  year={2019},
  publisher={Frontiers Media SA}
}

@article{resnet,
  title={Deep residual learning in spiking neural networks},
  author={Fang, Wei and Yu, Zhaofei and Chen, Yanqi and Huang, Tiejun and Masquelier, Timoth{\'e}e and Tian, Yonghong},
  journal={Advances in Neural Information Processing Systems},
  volume={34},
  pages={21056--21069},
  year={2021}
}
}

\clearpage
\setcounter{page}{1}
\maketitlesupplementary
\section{Centered Kernel Alignment (CKA)}
\label{Details of CKA}

In this work, we use Centered Kernel Alignment (CKA) as the similarity metric to measure the representation similarity between the source and target domains. Given two kernel (Gram) matrices \(K, L \in \mathbb{R}^{n \times n}\), CKA is defined as:
\begin{equation}
\text{CKA}(K, L) = \frac{\text{HSIC}(K, L)}{\sqrt{\text{HSIC}(K, K)\,\text{HSIC}(L, L)}},
\label{eq:cka}
\end{equation}
where HSIC stands for the Hilbert-Schmidt Independence Criterion.

The HSIC between \(K\) and \(L\) is computed as:
\begin{equation}
\text{HSIC}(K, L) = \frac{1}{(n - 1)^2} \, \text{tr}(K J L J),
\label{eq:hsic}
\end{equation}
where \(\text{tr}(\cdot)\) is the matrix trace, and \(J\) is the centering matrix defined as:
\begin{equation}
J = I_n - \frac{1}{n} \mathbf{1} \mathbf{1}^{\top},
\end{equation}
with \(I_n\) being the \(n \times n\) identity matrix.

In practice, we use the linear kernel \(k(x_i, x_j) = x_i^{\top} x_j\), and the kernel matrices become:
\begin{equation}
K = X X^{\top}, \quad L = Y Y^{\top},
\end{equation}
where \(X\) and \(Y\) are feature matrices derived from the penultimate layer's membrane potentials for the mixed and event streams, respectively.
The resulting CKA value is then used in the domain alignment loss to align the source and target domain features in a shared representation space.

\onecolumn
\section{Proof of Theorem 3.2 and Remark 3.3}
\label{proof}
We proceed in three steps: (i) compute the means of $G_{\mathrm{TSM}}$ and $G_{\mathrm{BM}}$; (ii) compute their covariances from one i.i.d.\ sample contribution using the time–index–invariant second–order structure in Assumption 3.1; (iii) compare the two covariances.

\paragraph{Step 1: Means.}
From Assumption 3.1, the per–sample averages are
\begin{equation}
\begin{split}
Z_i^{\mathrm{TSM}}
&=\frac{1}{T}\!\left(\sum_{t=1}^{n_a} g_a^{i,t}+\sum_{t=n_a+1}^{T} g_e^{i,t}\right),\\
Z_i^{\mathrm{BM}}
&=\begin{cases}
\frac{1}{T}\sum_{t=1}^{T} g_a^{i,t}, & C_i=a,\\[3pt]
\frac{1}{T}\sum_{t=1}^{T} g_e^{i,t}, & C_i=e,
\end{cases}
\end{split}\nonumber
\end{equation}
and $G_{\square}=\frac{1}{B}\sum_{i=1}^B Z_i^{\square}$, $\square\in\{\mathrm{TSM},\mathrm{BM}\}$.
Linearity of expectation with $\mathbb{E}[g_a^{i,t}]=\mu_a$, $\mathbb{E}[g_e^{i,t}]=\mu_e$ gives
\[
\mathbb{E}[Z_i^{\mathrm{TSM}}]
=\frac{n_a}{T}\mu_a+\frac{n_e}{T}\mu_e
=(1-\alpha)\mu_a+\alpha\mu_e.
\]
For BM, by the assignment $C_i$ with $\Pr[C_i=e]=\alpha$ and independence of $C_i$ from $\{g_\bullet^{i,t}\}$,
\[
\mathbb{E}[Z_i^{\mathrm{BM}}]=(1-\alpha)\mu_a+\alpha\mu_e.
\]
Averaging over $i$ preserves the mean, hence
\[
\mathbb{E}[G_{\mathrm{TSM}}]=\mathbb{E}[G_{\mathrm{BM}}]=(1-\alpha)\mu_a+\alpha\mu_e.
\]

\paragraph{Step 2: Covariances.}
Because samples are i.i.d., for either estimator
\[
\mathrm{Cov}(G_{\square})=\frac{1}{B}\,\mathrm{Cov}(Z_i^{\square}),\qquad \square\in\{\mathrm{TSM},\mathrm{BM}\}.
\]

\emph{TSM.} Define $S_a^i:=\sum_{t=1}^{n_a} g_a^{i,t}$ and $S_e^i:=\sum_{t=n_a+1}^{T} g_e^{i,t}$ so that $Z_i^{\mathrm{TSM}}=\tfrac{1}{T}(S_a^i+S_e^i)$.
Using time–index invariance (Assumption 3.1):
\[
\mathrm{Cov}(S_a^i)=\sum_{t=1}^{n_a}\!\mathrm{Var}(g_a^{i,t})+\!\!\sum_{\substack{t\neq s\\1\le t,s\le n_a}}\!\!\mathrm{Cov}(g_a^{i,t},g_a^{i,s})
=n_a\Sigma_a+n_a(n_a\!-\!1)R_a,
\]
\[
\mathrm{Cov}(S_e^i)=n_e\Sigma_e+n_e(n_e\!-\!1)R_e,
\qquad
\mathrm{Cov}(S_a^i,S_e^i)=\sum_{\substack{t\le n_a\\ n_a< s\le T}}\!\!\mathrm{Cov}(g_a^{i,t},g_e^{i,s})
=n_a n_e\,R_{ae}.
\]
Hence
\[
\mathrm{Cov}(Z_i^{\mathrm{TSM}})
=\frac{1}{T^2}\!\left[
n_a\Sigma_a+n_a(n_a\!-\!1)R_a+n_e\Sigma_e+n_e(n_e\!-\!1)R_e
+n_a n_e\,(R_{ae}+R_{ae}^{\top})
\right]\!.
\]
Substituting $n_a=(1-\alpha)T$, $n_e=\alpha T$ yields
\begin{equation}
\begin{split}
\mathrm{Cov}(Z_i^{\mathrm{TSM}})
&=\frac{1}{T}\Big[
(1-\alpha)\Sigma_a+\alpha\Sigma_e
+(1-\alpha)\big((1-\alpha)T-1\big)R_a
\\
&\hspace{15mm}
+\alpha\big(\alpha T-1\big)R_e
+(1-\alpha)(\alpha T)\big(R_{ae}+R_{ae}^{\top}\big)
\Big],
\end{split}\nonumber
\end{equation}
and therefore
\begin{equation}
\begin{split}
\mathrm{Cov}(G_{\mathrm{TSM}})
=\frac{1}{B\,T}\Big[
(1-\alpha)\Sigma_a+\alpha\Sigma_e
+(1-\alpha)\big((1-\alpha)T-1\big)R_a
\\
\hspace{17mm}
+\alpha\big(\alpha T-1\big)R_e
+(1-\alpha)(\alpha T)\big(R_{ae}+R_{ae}^{\top}\big)
\Big].
\end{split}\nonumber
\end{equation}

\emph{BM.} Condition on $C_i$ and apply the law of total covariance:
\[
\mathrm{Cov}(Z_i^{\mathrm{BM}})
=\mathbb{E}\!\left[\mathrm{Cov}(Z_i^{\mathrm{BM}}\mid C_i)\right]
+\mathrm{Cov}\!\left(\mathbb{E}[Z_i^{\mathrm{BM}}\mid C_i]\right).
\]
Given $C_i=a$, $Z_i^{\mathrm{BM}}=\tfrac{1}{T}\sum_{t=1}^{T} g_a^{i,t}$, so
\[
\mathrm{Cov}(Z_i^{\mathrm{BM}}\mid C_i=a)
=\frac{1}{T^2}\!\left[T\Sigma_a+T(T-1)R_a\right]
=\frac{1}{T}\Big(\Sigma_a+(T-1)R_a\Big),
\]
and symmetrically for $C_i=e$.
Averaging over $C_i$ with $\Pr[C_i=e]=\alpha$ gives the within–class contribution
\[
\frac{1}{T}\Big[(1-\alpha)\big(\Sigma_a+(T-1)R_a\big)
+\alpha\big(\Sigma_e+(T-1)R_e\big)\Big].
\]
For the between–class term,
$\mathbb{E}[Z_i^{\mathrm{BM}}\mid C_i=a]=\mu_a$, $\mathbb{E}[Z_i^{\mathrm{BM}}\mid C_i=e]=\mu_e$, hence
\[
\mathrm{Cov}\!\left(\mathbb{E}[Z_i^{\mathrm{BM}}\mid C_i]\right)
=\alpha(1-\alpha)\,(\mu_e-\mu_a)(\mu_e-\mu_a)^\top.
\]
Therefore,
\begin{equation}
\begin{split}
\mathrm{Cov}(Z_i^{\mathrm{BM}})
&=\frac{1}{T}\Big[(1-\alpha)\big(\Sigma_a+(T-1)R_a\big)
+\alpha\big(\Sigma_e+(T-1)R_e\big)\Big]
\\
&\hspace{17mm}
+\alpha(1-\alpha)\,(\mu_e-\mu_a)(\mu_e-\mu_a)^\top,
\end{split}\nonumber
\end{equation}
and thus
\begin{equation}
\begin{split}
\mathrm{Cov}(G_{\mathrm{BM}})
=\frac{1}{B\,T}\Big[(1-\alpha)\big(\Sigma_a+(T-1)R_a\big)
+\alpha\big(\Sigma_e+(T-1)R_e\big)\Big]
\\
+\frac{\alpha(1-\alpha)}{B}\,(\mu_e-\mu_a)(\mu_e-\mu_a)^\top.
\end{split}\nonumber
\end{equation}


\paragraph{Step 3: Difference.}
Subtracting the two covariances yields
\[
\mathrm{Cov}(G_{\mathrm{BM}})-\mathrm{Cov}(G_{\mathrm{TSM}})
=\frac{\alpha(1-\alpha)}{B}\Big[(\mu_e-\mu_a)(\mu_e-\mu_a)^\top
+ R_a + R_e - R_{ae}-R_{ae}^{\top}\Big].
\]
This RHS is PSD: $(\mu_e-\mu_a)(\mu_e-\mu_a)^\top\succeq0$, and by the block covariance
$\begin{bmatrix}R_a & R_{ae}\\ R_{ae}^\top & R_e\end{bmatrix}\succeq0$ we have
$R_a+R_e-R_{ae}-R_{ae}^\top\succeq0$ (take $w=[v;-v]$). Multiplication by $\alpha(1-\alpha)/B\ge0$ preserves PSD, so
$\mathrm{Cov}(G_{\mathrm{BM}})-\mathrm{Cov}(G_{\mathrm{TSM}})\succeq0$, i.e., $\mathrm{Cov}(G_{\mathrm{TSM}})\preceq \mathrm{Cov}(G_{\mathrm{BM}})$.

\end{document}